\documentclass[letterpaper, 10 pt, journal, twoside]{IEEEtran}  

\IEEEoverridecommandlockouts                              

\usepackage{graphics} 
\usepackage{epsfig} 
\usepackage{amsmath} 
\usepackage{amssymb}  
\usepackage{hyperref}
\usepackage{xcolor}
\usepackage{tikz}
\usetikzlibrary{calc}
\usetikzlibrary{arrows}
\usepackage{pgfplots}
\usepackage{siunitx}
\usepackage{adjustbox}
\usepackage{bm}
\usepackage{xcolor}
\usepackage{soul}
\usepackage[ruled,vlined,linesnumbered]{algorithm2e}
\usepackage{multirow}
\usepackage{booktabs}

\DeclareMathOperator*{\argminA}{arg\,min} 

\usepackage[caption=false,font=scriptsize]{subfig}
\usepackage{stfloats}

\pgfplotsset{compat=1.18}

\title{\LARGE \bf
DQ-NMPC: Dual-Quaternion NMPC for Quadrotor Flight
}
\begin{document}


\author{Luis F. Recalde$^{1}$, Dhruv Agrawal$^{1}$, Jon Arrizabalaga$^{2}$, and Guanrui Li$^{1}$
\thanks{Accepted to IEEE Robotics and Automation Letters.}
\thanks{
$^1$The authors are with the Department of Robotics Engineering, Worcester Polytechnic Institute, Worcester, MA 01609, USA. {\tt\footnotesize email: \{lfrecalde, dagrawal, gli7\}@wpi.edu}.}
\thanks{$^2$The author is with the Robotics Institute, Carnegie Mellon University, Pittsburgh, PA 15213, USA. {\tt\footnotesize email: jarrizab@andrew.cmu.edu}.}
\thanks{The authors acknowledge Amazon Robotics for their support on this paper.}
}
\maketitle
\begin{abstract}
 MAVs have great potential to assist humans in complex tasks, with applications ranging from logistics to emergency response. Their agility makes them ideal for operations in complex and dynamic environments. However, achieving precise control in agile flights remains a significant challenge, particularly due to the underactuated nature of quadrotors and the strong coupling between their translational and rotational dynamics.
In this work, we propose a novel NMPC framework based on dual-quaternions (DQ-NMPC) for quadrotor flight. By representing both quadrotor dynamics and the pose error directly on the dual-quaternion manifold, our approach enables a compact and globally non-singular formulation that captures the quadrotor coupled dynamics. We validate our approach through simulations and real-world experiments, demonstrating better numerical conditioning and significantly improved tracking performance, with reductions in position and orientation errors of up to $56.11\%$ and $56.77\%$, compared to a conventional baseline NMPC method. Furthermore, our controller successfully handles aggressive trajectories, reaching maximum speeds up to $13.66 ~\si{m/s}$ and accelerations reaching $4.2~\si{g}$ within confined space conditions of dimensions $11\si{m} \times 4.5\si{m} \times 3.65\si{m}$ under which the baseline controller fails.

\end{abstract}



\section*{Supplementary material}
Code:
\url{https://github.com/acp-lab/dq_nmpc.git}

Video:
\url{https://youtu.be/3XxvbsIoenM}

Page:
\url{https://acp-lab.github.io/dq-nmpc.github.io/}

\section{Introduction}
\label{sec:Introduction}
\IEEEPARstart{M}{icro} Aerial Vehicles (MAVs) have great potential in assisting humans in complex tasks such as environmental monitoring, logistics, and emergency response. Their agility and versatility make them ideal for operations in complex and dynamic environments.  
Nonlinear Model Predictive Control (NMPC) has been widely demonstrated to be effective in enabling agile maneuvers for MAVs~\cite{9794477, s22134712}.
However, these approaches are often computationally intensive, especially compared to classical methods, such as geometric control \cite{9121690}. Moreover, most existing NMPC frameworks treat translational and rotational dynamics separately, neglecting the inherent coupling between these states and potentially limiting performance in agile flight scenarios~\cite{10049101, 9720967}. These methods typically rely on a time-scale separation between rotational and translational dynamics. However, insufficient convergence in the rotational dynamics can severely impact the overall stability and performance of the system \cite{5717652}.

Researchers have also investigated coupled representations of rigid body dynamics in the Special Euclidean group $SE(3)$ for MPC. MPC formulations based on error-state linearization in $SE(3)$ have been proposed in \cite{9917382, 9981282}. Although these approaches capture coupled dynamics, their reliance on local error approximations limits their validity to small deviations from a reference trajectory. In contrast, other methods operate directly on $SE(3)$ without relying on linearization, often computing rotations using over-parameterized representations from the special orthogonal group $SO(3)$. Although this avoids approximation, it introduces a non-minimal state representation increasing computational complexity \cite{Pereira2021, 7320769}.

\begin{figure}[t]
    \centering
\includegraphics[width=1.0\columnwidth]{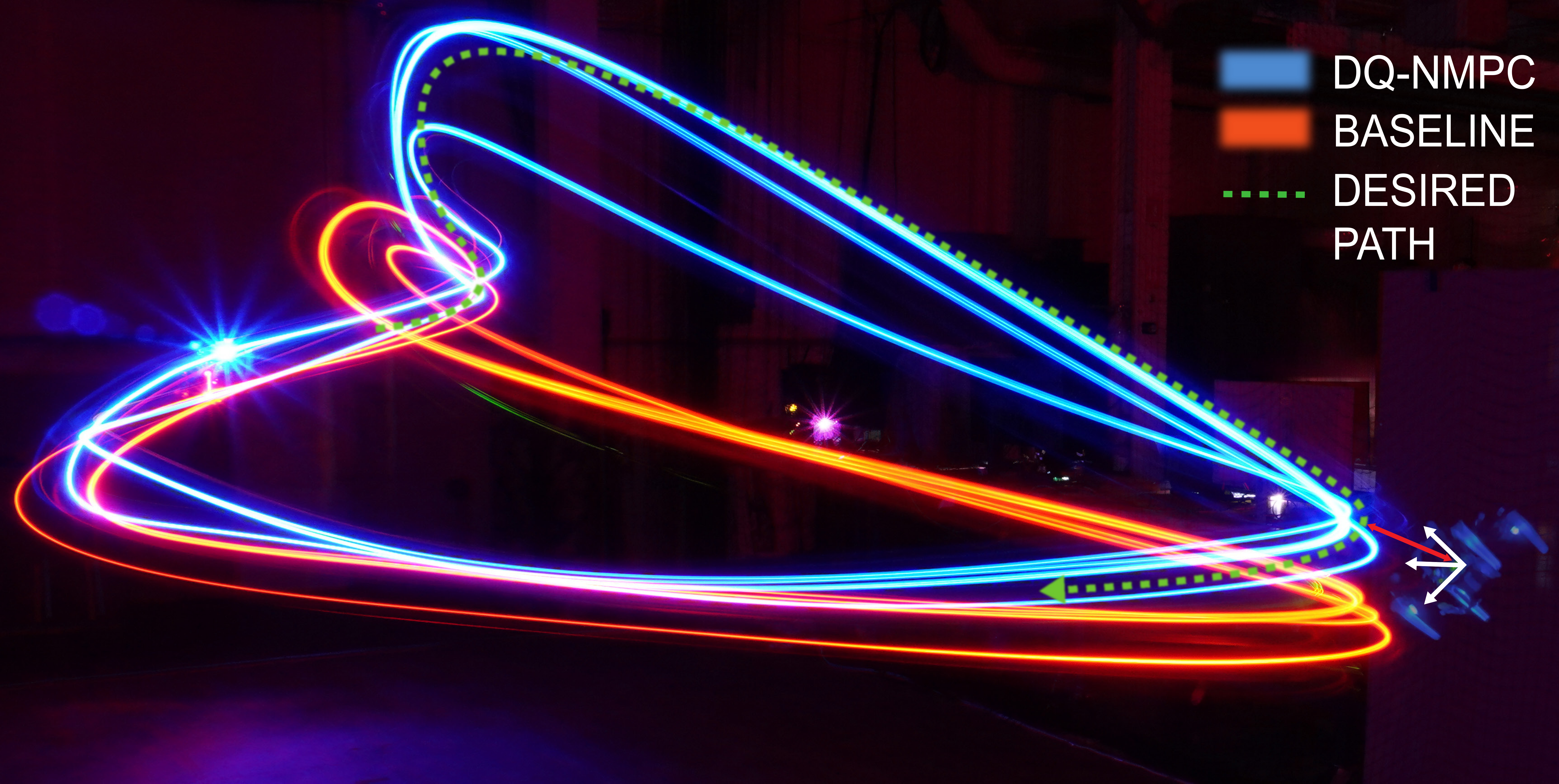}
    \caption{Trajectory tracking experiment comparing Dual-Quaternion NMPC and Baseline NMPC. Green lines indicate the desired trajectory, blue lines show the system's evolution under the DQ-NMPC controller, and orange lines represent the results obtained with the baseline NMPC. 
    }
    \label{fig:real_system}
    \vspace{-15pt}
\end{figure}

In contrast, unit dual-quaternions offer a more compact and globally non-singular representation of $SE(3)$ for rigid body dynamics, using only 8 parameters compared to the 12 required by $SE(3)$ \cite{662755, 6669564}. Moreover, dual-quaternion algebra provides a unified framework to represent twists, wrenches, and various geometric primitives, which makes it suitable for robotic modeling and control tasks \cite{FIGUEREDO2021109817, arrizabalaga2023pose}. 
Despite these advantages, relatively few studies have addressed control problems using unit dual-quaternions \cite{Constrained, arrizabalaga2023pose}, highlighting the need for further exploration.

We propose an NMPC framework for quadrotor flight that employs dual quaternions to achieve a unified and compact representation of the system dynamics. In addition, the proposed approach leverages a quadratic cost function in the Lie algebra of dual-quaternions that simultaneously encapsulates both rotation and translation errors.

In summary, our contributions are the following:
\begin{itemize}
  
\item We introduce a novel NMPC framework formulated directly on the dual-quaternion manifold, offering a unified and compact representation of quadrotor dynamics. The framework employs a quadratic cost function defined in the Lie algebra, leveraging a unified pose error that couples translation and rotation, enabling consistently scaled cost with respect to pose errors.

\item We validate the proposed control method through various sets of simulation experiments, showing that the proposed method improves convergence of the optimization problem and enables faster convergence to the desired state compared to the baseline NMPC method.

\item We test our method in real-world experiments, demonstrating that DQ-NMPC reduces position and orientation tracking errors by up to $56.11\%$ and $56.77\%$, compared to the baseline for velocities of up to $7.5~\si{m/s}$. In addition, our controller can track
aggressive trajectories with maximum velocities up to $13.66~\si{m/s}$ and accelerations up to $4.2~\si{g}$ in confined space conditions $11\si{m} \times 4.5\si{m} \times 3.65\si{m}$, where the baseline fails.
\end{itemize}

This work is organized as follows: Section \ref{System Dynamics} introduces the dual-quaternion preliminaries and system dynamics modeling using dual-quaternions. Section \ref{Control Formulation} presents the proposed DQ-NMPC. Section \ref{Experimental Results} reports the results of the simulation and real-world experiments, and Section \ref{Conclusion} presents the conclusions and future research directions.

\section{System Dynamics}\label{System Dynamics}
In this section, we introduce the compact representation of the system dynamics using dual-quaternions. We first present brief mathematical preliminaries on dual quaternions, followed by a formulation of quadrotor dynamics using dual quaternions, which enables a unified pose representation.
\subsection{Mathematical Preliminaries}
This section introduces dual-quaternion algebra, briefly revisiting the fundamental concepts of quaternions and dual numbers as prerequisites. Additional details on these topics can be found in \cite{8742769, arrizabalaga2023pose, WANG2013225}.

\subsubsection{Quaternions} 
Quaternions are an extension of complex numbers with a compact set defined as:
$$
 \mathbb{H} \triangleq \{ q_1 + \hat{i} q_2  + \hat{j} q_3  + \hat{k} q_4 : q_1, q_2, q_3, q_4 \in \mathbb{R}\}, 
$$
where the imaginary units
$\hat{i}$, $\hat{j}$, and $\hat{k}$ have the following properties $\hat{i}^2 = \hat{j}^2 = \hat{k}^2 = \hat{i} \hat{j} \hat{k} = -1$. 
A quaternion can be expressed as $\mathbf{q}=\textrm{Re}(\mathbf{q}) + \textrm{Im}(\mathbf{q})$, where $\textrm{Re} (\mathbf{q})$ represents the real part, and $\textrm{Im}(\mathbf{q})$ is the imaginary part that contains $\{\hat{i}, \hat{j}, \hat{k}\}$. The norm of quaternions can be formulated as $\| \mathbf{q} \| = \sqrt{\mathbf{q} \otimes \mathbf{q}^{*} }$, where $\mathbf{q}^{*} = \textrm{Re}(\mathbf{q}) - \textrm{Im}(\mathbf{q})$ is the conjugate of $\mathbf{q}$  and $\otimes$ is the quaternion multiplication operator.

We can represent three-dimensional vectors as a subset of quaternions containing only imaginary components, known as \textit{pure quaternions}. This subset is defined as
$$
\mathbb{H}_p \triangleq \{ \mathbf{q} \in \mathbb{H} : \textrm{Re}(\mathbf{q}) = 0 \}\,.
$$

We can also represent a rigid body's orientation by \textit{unit quaternions}, whose norm is $1$, defined as
$$
\mathbb{S}^{3} \triangleq \{ \mathbf{q} \in \mathbb{H} : \| \mathbf{q} \| = 1\}\,.
$$
\subsubsection{Dual Numbers}
To define dual-quaternions, it is essential to first introduce a new algebraic unit known as the dual unit $\varepsilon$. The algebra generated by $\varepsilon$ was first introduced by Clifford \cite{cliffort} and is characterized by the properties $\varepsilon \neq 0$ and $\varepsilon^2 = 0$. A dual number can be expressed as $\hat{\mathbf{a}} = \mathbf{a} + \varepsilon \mathbf{b}$, where $\mathbf{a}$ is the primary part and $\mathbf{b}$ is the dual part. These components can be extracted using the operators $\mathcal{P}(\hat{\mathbf{a}})$ and $\mathcal{D}(\hat{\mathbf{a}})$, respectively, such that $\hat{\mathbf{a}} = \mathcal{P}(\hat{\mathbf{a}}) +  \varepsilon \mathcal{D}(\hat{\mathbf{a}})\,$.

\subsubsection{Dual quaternions} Dual-quaternions are dual numbers in which both the primary and dual parts are quaternions. We can formulate the set of dual-quaternions as follows
$$
\mathcal{H} \triangleq \{ \mathbf{q}_p + \epsilon \mathbf{q}_d: (\mathbf{q}_p, \mathbf{q}_d) \in \mathbb{H}, \epsilon^2 = 0, \epsilon \neq 0 \}\,,
$$
thus, given the dual-quaternion $\hat{\mathbf{q}} =  a + \hat{i} b + \hat{j} c + \hat{k} d + \epsilon( e + \hat{i} f + \hat{j} g + \hat{k} h)$, we can establish the following operations
$$
\mathcal{P}(\hat{\mathbf{q}}) = a + \hat{i} b + \hat{j} c + \hat{k} d~~~~\mathcal{D}(\hat{\mathbf{q}}) = e + \hat{i} f + \hat{j} g + \hat{k} h\,.
$$
$$
\textrm{Re}(\hat{\mathbf{q}}) = a + \epsilon( e ) ~~~ \textrm{Im}(\hat{\mathbf{q}}) = \hat{i} b + \hat{j} c + \hat{k} d + \epsilon (\hat{i} f + \hat{j} g + \hat{k} h)
$$

We can define the full conjugate of dual-quaternions as \(\hat{\mathbf{q}}^{*} = \textrm{Re}(\hat{\mathbf{q}}) - \textrm{Im}(\hat{\mathbf{q}})\) and the norm as \(\| \hat{\mathbf{q}} \| = \sqrt{\hat{\mathbf{q}} \boxtimes \hat{\mathbf{q}}^{*} } \), where \(\boxtimes\) is the dual-quaternion multiplication operator. With the definition of the norm of dual quaternion, we can further define the unit dual quaternions as follows, 
$$
\mathcal{S} \triangleq \{ \hat{\mathbf{q}} \in \mathcal{H}:  \| \hat{\mathbf{q}} \| = 1\}.
$$
which we will leverage later in Section \ref{Quadrotor Dynamics} to represent the pose of the quadrotor in a compact fashion. In addition, we further introduce \emph{pure dual-quaternions}, defined as
$$
\mathcal{H}_p \triangleq \{ \hat{\mathbf{q}} \in \mathcal{H}: \textrm{Re}(\hat{\mathbf{q}}) = 0 \},$$
which we will use to represent the twist and wrench of a rigid body in Section \ref{Quadrotor Dynamics}.

\subsection{Classical Quadrotor Dynamics}
\begin{figure}[t]
    \centering
\includegraphics[width=0.8\columnwidth]{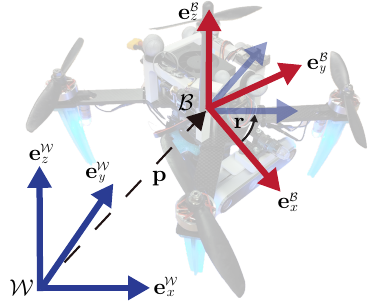}
    \caption{Aerial system representation with  inertial $\mathcal{W}$ and body frames $\mathcal{B}$}
    \label{fig:system_frames}
    \vspace{-15pt}
\end{figure}

We define the inertial frame as $\mathcal{W} = \{ \mathbf{e}^{\mathcal{W}}_x, \mathbf{e}^{\mathcal{W}}_y, \mathbf{e}^{\mathcal{W}}_z \}$, and the body frame as $\mathcal{B} = \{ \mathbf{e}^{\mathcal{B}}_x, \mathbf{e}^{\mathcal{B}}_y, \mathbf{e}^{\mathcal{B}}_z \}$, which is attached to the quadrotor’s center of mass (CoM). The unit quaternion $\mathbf{r} \in \mathbb{S}^3$ represents the orientation of the body frame $\mathcal{B}$ with respect to the inertial frame $\mathcal{W}$, as shown in Fig.~\ref{fig:system_frames}.

The classical quadrotor dynamics is decoupled into translation and rotation motion, which is widely used in research literature~\cite{9794477}, as shown in the following:
\begin{align}
\dot{\mathbf{p}} & =  \mathbf{v}, \dot{\mathbf{r}} =  \frac{1}{2}  \mathbf{r} \otimes {\bm{\omega}}\\
\dot{\mathbf{v}} &= \frac{f}{m} \mathbf{r} \otimes \mathbf{e}^{\mathcal{W}}_z \otimes \mathbf{r}^{*} - \mathbf{g},\label{eq:1.3}\\
\dot{\bm{\omega}} &= \mathbf{J}^{-1}(\bm{\tau} - \bm{\omega} \times (\mathbf{J} \bm{\omega}))\label{eq:1.4}, 
\end{align}
where $(\mathbf{p}, \mathbf{v}) \in \mathbb{R}^{3}$ denote the position and velocity of the quadrotor with respect to the inertial frame $\mathcal{W}$, $\bm{\omega} \in \mathbb{R}^{3}$ is the quadrotor's angular velocity in the body frame $\mathcal{B}$. $\mathbf{g} = g \mathbf{e}^{\mathcal{W}}_z$ is the gravity vector, where $g = 9.81 \, \mathrm{m/s^2}$ and $\mathbf{e}^{\mathcal{W}}_z = \begin{bmatrix} 0 & 0 & 1 \end{bmatrix}^T$. The inertia matrix of the aerial vehicle is given by $\mathbf{J} = \mathrm{diag}(J_{xx},~J_{yy},~J_{zz}) \in \mathbb{R}^{3 \times 3}$, and its mass is denoted by $m \in \mathbb{R}_{> 0}$. Finally, the control inputs consist of the thrust force $f \in \mathbb{R}_{> 0}$ and the torque vector $\bm{\tau} \in \mathbb{R}^3$ expressed in the body frame $\mathcal{B}$.

\subsection{Quadrotor Dynamics Based on Dual-Quaternions}\label{Quadrotor Dynamics}
In this section, we present the quadrotor dynamics modeling using dual-quaternions.
First of all, a unit dual-quaternion $\hat{\mathbf{q}} \in \mathcal{S}$ can represent a rigid-body transformation consisting of a translation $\mathbf{p} \in \mathbb{H}_p$ followed by a rotation $\mathbf{r} \in \mathbb{S}^{3}$, shown as follows:
\begin{equation} \label{eq:3}
\hat{\mathbf{q}} = \mathbf{r} + \frac{1}{2} \varepsilon \left( \mathbf{p} \otimes \mathbf{r} \right),
\end{equation}
where the position $\mathbf{p}$ can be explicitly recovered through $\mathbf{p} = 2 \mathcal{D}(\hat{\mathbf{q}}) \otimes \mathcal{P}(\hat{\mathbf{q}})^{*}$.
Taking the time derivative of Eq.~\eqref{eq:3} yields the kinematics of the pose:
\begin{align}
\label{eq:4.1}
\dot{\hat{\mathbf{q}}} &= \frac{1}{2} \hat{\mathbf{q}}  \boxtimes \hat{\bm{\omega}},\\
\label{eq:4.2}
\hat{\bm{\omega}} &= \bm{\omega} + \epsilon (\mathbf{r}^{*} \otimes {\mathbf{v}} \otimes \mathbf{r}),
\end{align}
where $\hat{\bm{\omega}} \in \mathcal{H}_p$ denotes the dual twist expressed in the body frame $\mathcal{B}$, $(\bm{\omega}, \mathbf{v}) \in \mathbb{H}_p$ represents the angular and  linear velocities as pure quaternions.
To express the quadrotor’s dynamics using  dual-quaternions, we take the time derivative of Equation~\eqref{eq:4.2}, yielding the following expression:
\begin{equation} \label{eq:5}
 \begin{aligned}
  \dot{\hat{\bm{\omega}}} =  \dot{\bm{\omega}} + \epsilon((\textrm{ad}_{(\mathbf{r}^{*})} \dot{\mathbf{p}})\times \bm{\omega} +  \mathbf{r}^{*} \otimes \dot{\mathbf{v}} \otimes \mathbf{r}),
     \end{aligned}
\end{equation}
where $\textrm{ad}_{(\mathbf{r}^{*})} \dot{\mathbf{p}} = \mathbf{r}^{*} \otimes \dot{\mathbf{p}} \otimes \mathbf{r}$ is the adjoint transformation and the cross product operation $\times:$ between pure quaternions is expressed as follows:
$$
\textrm{ad}_{(\mathbf{r}^{*})} \dot{\mathbf{p}}\times \bm{\omega} = \frac{\textrm{ad}_{(\mathbf{r}^{*})} \dot{\mathbf{p}} \otimes \bm{\omega} - \bm{\omega} \otimes \textrm{ad}_{(\mathbf{r}^{*})} \dot{\mathbf{p}}}{2}\,.
$$

By substituting (\ref{eq:1.3}) and (\ref{eq:1.4}) into (\ref{eq:5}), we have
\begin{equation} \label{eq:6}
 \begin{aligned}
   \dot{\hat{\bm{\omega}}} =  \hat{\mathbf{f}} + \hat{\mathbf{u}}\,,
     \end{aligned}
\end{equation}
where
\begin{subequations}
\begin{align}
&\hat{\mathbf{f}} = \mathbf{a} + \epsilon((\textrm{ad}_{(\mathbf{r}^{*})} \dot{\mathbf{p}})\times \bm{\omega} - (\mathbf{r}^{*} \otimes \mathbf{g} \otimes \mathbf{r}))\,,\notag\\
&\hat{\mathbf{u}} =  \mathbf{J}^{-1} \bm\tau + \epsilon (\frac{f}{m} \mathbf{e}^{\mathcal{W}}_z),\,\,\,\notag \mathbf{a} = -\mathbf{J}^{-1} \bm{\omega} \times \mathbf{J} \bm{\omega}\,.\notag
\end{align}
\end{subequations}
Hence, based on eqs.~(\ref{eq:4.1}) and (\ref{eq:6}), the quadrotor's dynamics can be compactly written using dual-quaternions as follows:
\begin{equation}\label{eq:7}
\dot{\hat{\mathbf{x}}} = \mathbf{f}_d(\hat{\mathbf{x}}, \hat{\mathbf{u}})= \begin{bmatrix} \frac{1}{2} \hat{\mathbf{q}}  \boxtimes \hat{\bm{\omega}} \\ \hat{\mathbf{f}} + \hat{\mathbf{u}}\,
\end{bmatrix},
\end{equation} 
where $\hat{\mathbf{x}} = \begin{bmatrix} \hat{\mathbf{q}} & \hat{\bm{\omega}} \end{bmatrix}^T \in \mathcal{S} \times \mathcal{H}_p$ denotes the internal state, comprising the unit dual-quaternion $\hat{\mathbf{q}}$ and the dual twist $\hat{\bm{\omega}}$; the control input is represented by  $\hat{\mathbf{u}} \in \mathcal{H}_p $. As shown in eq.~\eqref{eq:7}, we can represent the system’s pose and twist in a unified and compact form by leveraging dual-quaternions.

\section{DQ-NMPC}\label{Control Formulation}

\begin{figure*}
\centering
    \includegraphics[width=0.95\textwidth]{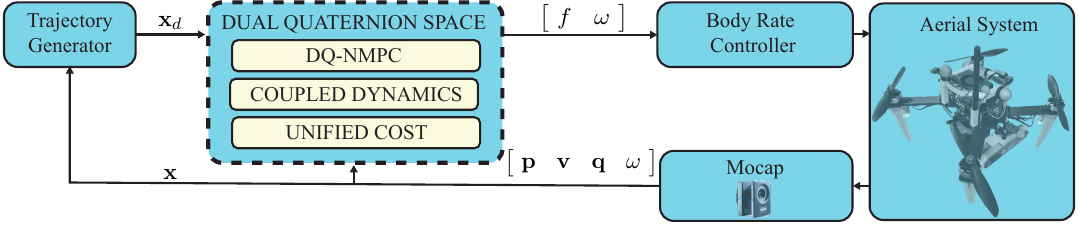}
    \caption{Control block diagram of the proposed DQ-NMPC, illustrating the flow of information through the DQ-NMPC formulation. $\mathbf{x}$ and $\mathbf{x}_d$ represent the current and desired pose of the quadrotor based on decoupled dynamics. These values are transformed into the dual-quaternion space using eqs.~\eqref{eq:3} and \eqref{eq:4.2} and subsequently used in the DQ-NMPC method.
    \label{fig:sys_scheme}}
    \vspace{-15pt}
\end{figure*}

This section introduces \textbf{D}ual-\textbf{Q}uaternion \textbf{N}onlinear \textbf{M}odel \textbf{P}redictive \textbf{C}ontrol \textbf{(DQ-NMPC)} method, an MPC framework developed for quadrotor flight based on dual-quaternions, as illustrated in Fig.~\ref{fig:sys_scheme}. The approach defines the evolution of quadrotor dynamics on a dual-quaternion manifold, which unifies translational and rotational dynamics by leveraging unit dual-quaternions and dual twists. The cost function of the proposed \textbf{DQ-NMPC} is designed to jointly penalize translational and rotational errors in a unified manner, using the left-invariant pose error and its projection onto the Lie algebra of dual-quaternions.

\subsection{Dual-Quaternion NMPC}

 NMPC formulations aim to compute a sequence of system states and control actions that optimize an objective function over a fixed time horizon, while explicitly incorporating the system dynamics and constraints into the optimal control problem. A general formulation of an NMPC can be written as follows:
 \begin{equation}\label{eq:nmpc}
\begin{split}
     \argminA_{\begin{matrix} \hat{\mathbf{x}}_1, \hdots, \hat{\mathbf{x}}_N, \\
         \hat{\mathbf{u}}_1, \hdots, \hat{\mathbf{u}}_{N-1}
     \end{matrix}}  \quad &  l_N(\hat{\mathbf{x}}_N) + \sum_{k=1}^{N-1} l(\hat{\mathbf{x}}_k, \hat{\mathbf{u}}_{k}) \\  
    \textrm{subject to: } & {\hat{\mathbf{x}}}_{k+1} =  \mathbf{f}(\hat{\mathbf{x}}_k,\hat{\mathbf{u}}_k), \forall k = 1,..., N-1\\
    &\mathbf{g}( {\hat{\mathbf{x}}}_k, \hat{\mathbf{u}}_k) \leq 0~~~~~~~~~~~~~~~~~~~~
\end{split}
\end{equation}
where $l_N(\hat{\mathbf{x}}_N)$ and $l(\hat{\mathbf{x}}, \hat{\mathbf{u}})$ represent the terminal and running cost functions, respectively. The dynamics of the discrete non-linear system is given by ${\hat{\mathbf{x}}}_{k+1} =  \mathbf{f}(\hat{\mathbf{x}}_k,\hat{\mathbf{u}}_k)$, and the state and input constraints are denoted by $\mathbf{g}( {\hat{\mathbf{x}}}, \hat{\mathbf{u}})$.

In the following, we present the detailed design of the proposed \textbf{DQ-NMPC} method for quadrotor flight. It includes 1) a cost function that simultaneously accounts for both translational and rotational errors by using a dual quaternion-based \textit{ left-invariant error} metric; 2) dynamics equations represented using unit dual-quaternions and dual twists, enabling a unified treatment of translation and orientation within the control scheme.

\subsubsection{Cost Function}
To enable a quadrotor to track the desired trajectory, we define the error between the current pose $\hat{\mathbf{q}}$ and the desired pose $\hat{\mathbf{q}}_d$ using the \textit{left-invariant error} as following:
\begin{equation}\label{eq:error}
{\hat{\mathbf{q}}}_e = \hat{\mathbf{q}}^*_d \boxtimes \hat{\mathbf{q}}\,,
\end{equation} 
\begin{figure}[t]
    \centering
\includegraphics[width=1.0\columnwidth]{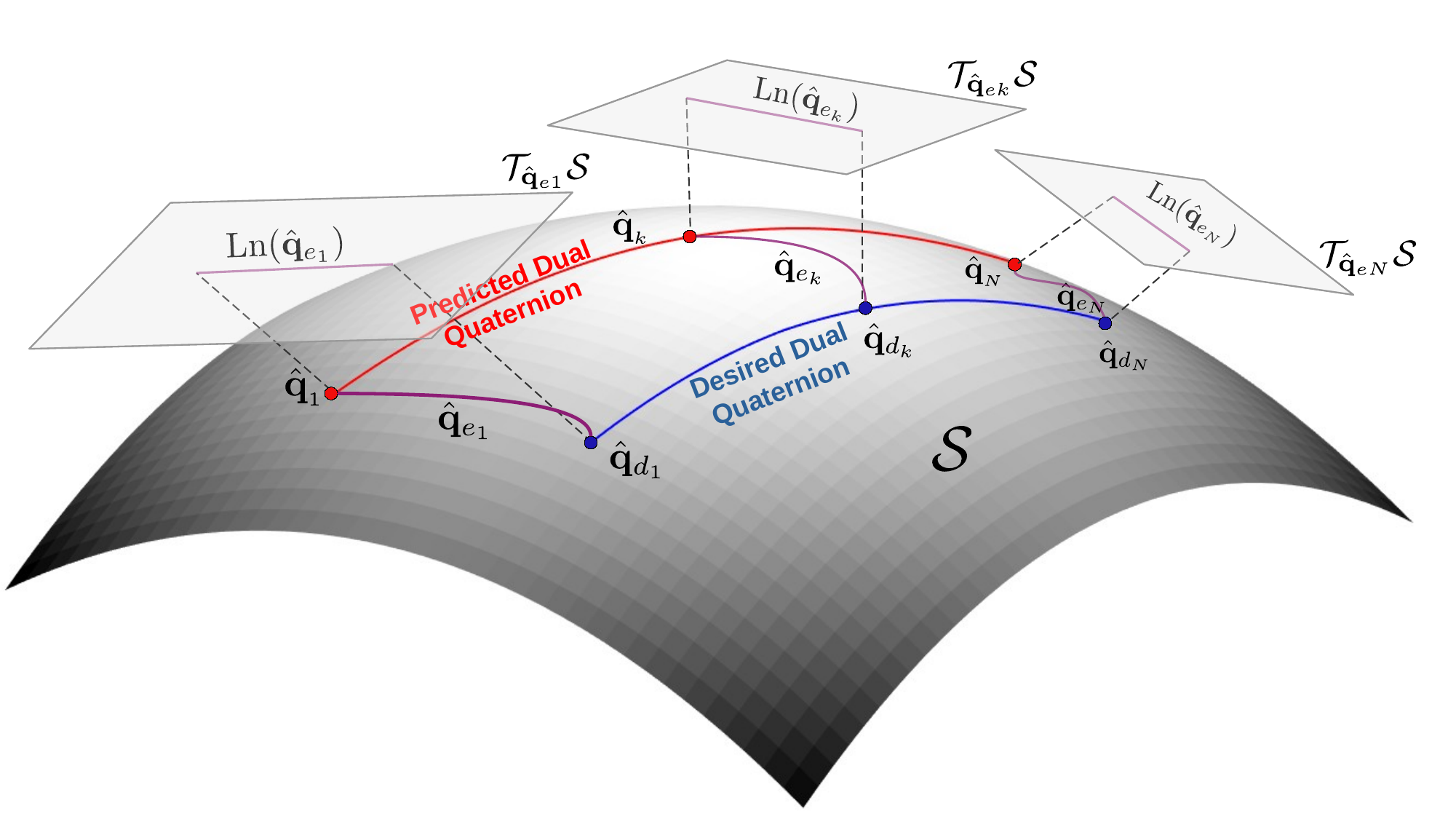}
    \caption{Interpretation of the DQ-NMPC formulation: Both the predicted and desired poses of the quadrotor are represented as unit dual-quaternions, and their difference is captured using a left-invariant error. This error lies on the dual-quaternion manifold and is projected onto the Lie algebra and included into the NMPC cost function.}
    \label{fig:DQ_NMPC}
\end{figure}
Additionally, the dual twist error can be expressed as:
 \begin{equation}\label{eq:error_twist}
\hat{\bm{\omega}}_{e} = \hat{\bm{\omega}}_{d} -\hat{\bm{\omega}},
\end{equation} 
 where $\hat{\bm{\omega}}_{d} \in \mathcal{H}_p$  represents the desired dual twist.
Using the errors above, we can further design the DQ-NMPC running and terminal cost functions that jointly penalize both pose and twist errors over a finite horizon $N$:
 \begin{equation}\label{eq:nmpc_dual}
\begin{split}
l(\hat{\mathbf{q}}_{eN}, \hat{\mathbf{\bm{\omega}}}_{eN}) + \sum_{k=1}^{N-1} l(\hat{\mathbf{q}}_{ek}, \hat{\mathbf{\bm{\omega}}}_{ek}) + l_{u}(\hat{\mathbf{u}}_{ek})\,,
\end{split}
\end{equation}
with
\begin{equation}
  l(\hat{\mathbf{q}}_e) = \Vert \textrm{Ln}(\hat{\mathbf{q}}_e) \Vert^{2}_{\mathbf{Q}_p} + \Vert\hat{\boldsymbol{\omega}}_{e}\Vert^{2}_{\mathbf{Q}_v},\,\,\, l_{u}(\hat{\mathbf{u}}_e) = \Vert  \hat{\mathbf{u}}_e\Vert^{2}_{\mathbf{R}}.
\end{equation}
The logarithmic map $\mathrm{Ln}(\hat{\mathbf{q}}) = \frac{1}{2} \left( \bm{\phi} + \varepsilon~ \mathbf{t} \right)$ extracts the minimal representation of pose error in the Lie algebra of dual-quaternions, where $\bm{\phi}$ and $\mathbf{t}$ represent the rotational and translational components. 
$\mathbf{Q}_p$ and $\mathbf{Q}_v$ are positive definite weighting matrices. Finally, $l_u$ penalizes the error from desired control actions $\hat{\mathbf{u}}_e = \hat{\mathbf{u}}_d - \hat{\mathbf{u}}$.
We choose this cost function inspired by \cite{9993143}, which showed that a proper left-invariant metric defined in the Lie algebra can ensure globally exponential convergence.

To provide further insights into the left-invariant error and its projection onto the Lie algebra used in our DQ-NMPC framework, we present a comparison of the orientation errors used by the DQ-NMPC and the baseline NMPC \cite{9794477}.
\begin{figure}[h]
    \centering
\includegraphics[width=1.0\columnwidth]{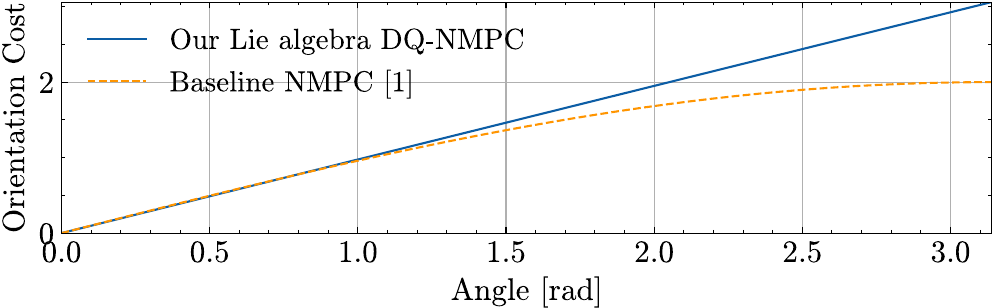}
    \caption{Comparison of different orientation error cost functions. The baseline cost function lacks
sensitivity to large orientation errors, which limits its performance and results in slower convergence}
    \label{fig:comparison}
\end{figure}
As shown in Fig. \ref{fig:comparison}, the consistent scaling of the error metric in DQ-NMPC can result in improved trajectory tracking accuracy, particularly during aggressive maneuvers that induce significant rotational errors. On the other hand, the baseline cost function remains nearly flat in the presence of large orientation errors, indicating a lack of sensitivity. This results in insufficient corrective control actions, which negatively affect the stability and tracking performance of the system.

\subsubsection{System Dynamics}
We integrate both translational and rotational dynamics employing unit dual-quaternions and dual twists. Hence, we apply a discrete form of eq.~\eqref{eq:7}, constructed with the fourth-order Runge-Kutta method \cite{8392463}, which enables efficient and singularity-free numerical integration of quaternions. This can be expressed as
\begin{equation}\label{eq:9}
   {\hat{\mathbf{x}}}_{k+1} = \mathbf{f}_{RK4d}({\mathbf{f}_d}(\hat{\mathbf{x}}_k,\hat{\mathbf{u}}_k), dt)\,,
\end{equation} 
where $dt$ denotes the sampling interval.

Fig.~\ref{fig:DQ_NMPC} provides a visual representation of the role of dual-quaternions within the NMPC framework.
\begin{figure*}[t]
    \centering
\includegraphics[width=\textwidth]{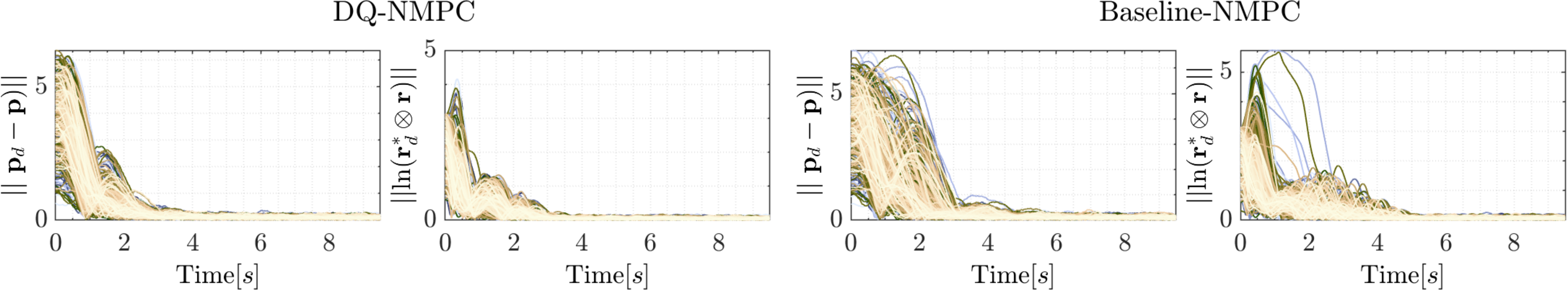}
    \caption{The position and orientation error metrics of both NMPC methods are evaluated in simulations using 600 randomly sampled initial poses for pose regulation. The initial translation vectors are sampled within the range $\begin{bmatrix} -4 & -4 & 0 \end{bmatrix}^{T}$ to $\begin{bmatrix} 4 & 4 & 4 \end{bmatrix}^{T}$, and the initial orientations are randomly sampled such that $\ln(\mathbf{r})$ is uniform in the interval $[0, \pi]$.}
    \label{fig:all_simulation_pose}
\end{figure*}

\subsubsection{Constraints}
To address potential issues related to the primary part of unit dual-quaternions, we introduce a non-linear constraint to enforce its unit norm. In addition, control input constraints are imposed to ensure feasibility within predefined bounds. These conditions can be compactly expressed as:
\begin{equation}\label{eq:17}
  \mathbf{g}_d(\hat{\mathbf{x}}, \hat{\mathbf{u}}) \leq 0,\,\,\,  
\mathbf{g}_d(\hat{\mathbf{x}}, \hat{\mathbf{u}}):= \begin{bmatrix}
    \|\mathcal{P}(\hat{\mathbf{q}})\| - 1  \\ 1-\|\mathcal{P}(\hat{\mathbf{q}})\| \\
    \hat{\mathbf{u}} -  \hat{\mathbf{u}}_{max} \\
    -\hat{\mathbf{u}} +  \hat{\mathbf{u}}_{min}
\end{bmatrix}
\end{equation} 
where $\hat{\mathbf{u}}_{max}$ and $\hat{\mathbf{u}}_{min}$ represent the maximum and minimum control actions.

\section{Experimental Results}\label{Experimental Results}

In this section, we present the experimental results of our proposed method in both simulation and real-world environments. We compared the proposed method with the recent state-of-the-art NMPC method in \cite{9794477}, which decoupled position and orientation in dynamics and cost functions. Both methods generate thrust and angular rate commands to the low-level angular rate controller.

In the following, we first present the simulation results, which demonstrate improved optimality, enhanced pose tracking performance, and robustness  compared to the baseline NMPC. Next, we present the results of the real world experiments, demonstrating that the proposed DQ-NMPC outperforms the baseline NMPC considering trajectory tracking tasks at velocities of up to $13.66~\si{m/s}$ and accelerations up to $4.2~\si{g}$.

\subsection{Simulation Results}\label{Simulation Results}

We conducted simulation experiments in two different scenarios, which led to a systematic analysis of the effectiveness of our proposed method.
\begin{itemize}
\item \textit{Pose Regulation:} This scenario focuses on a regulation problem using a well-identified model. We analyze the controllers' behavior in terms of pose error, optimality conditions, and convergence achieved by each control method.

\item \textit{Robustness Study:} To assess robustness, we introduce model mismatches and external disturbances during the simulation experiments. The performance of both controllers is evaluated across a range of trajectories under modeled mismatches and disturbances.
\end{itemize}
Our results show that the DQ-NMPC consistently achieves better convergence to the desired pose under different initial conditions. This formulation also yields better-conditioned solutions and requires fewer iterations within the optimization framework. Finally, it demonstrates improved performance under model mismatches compared to baseline.

\begin{figure}[t]
    \centering
\includegraphics[width=1.0\columnwidth]{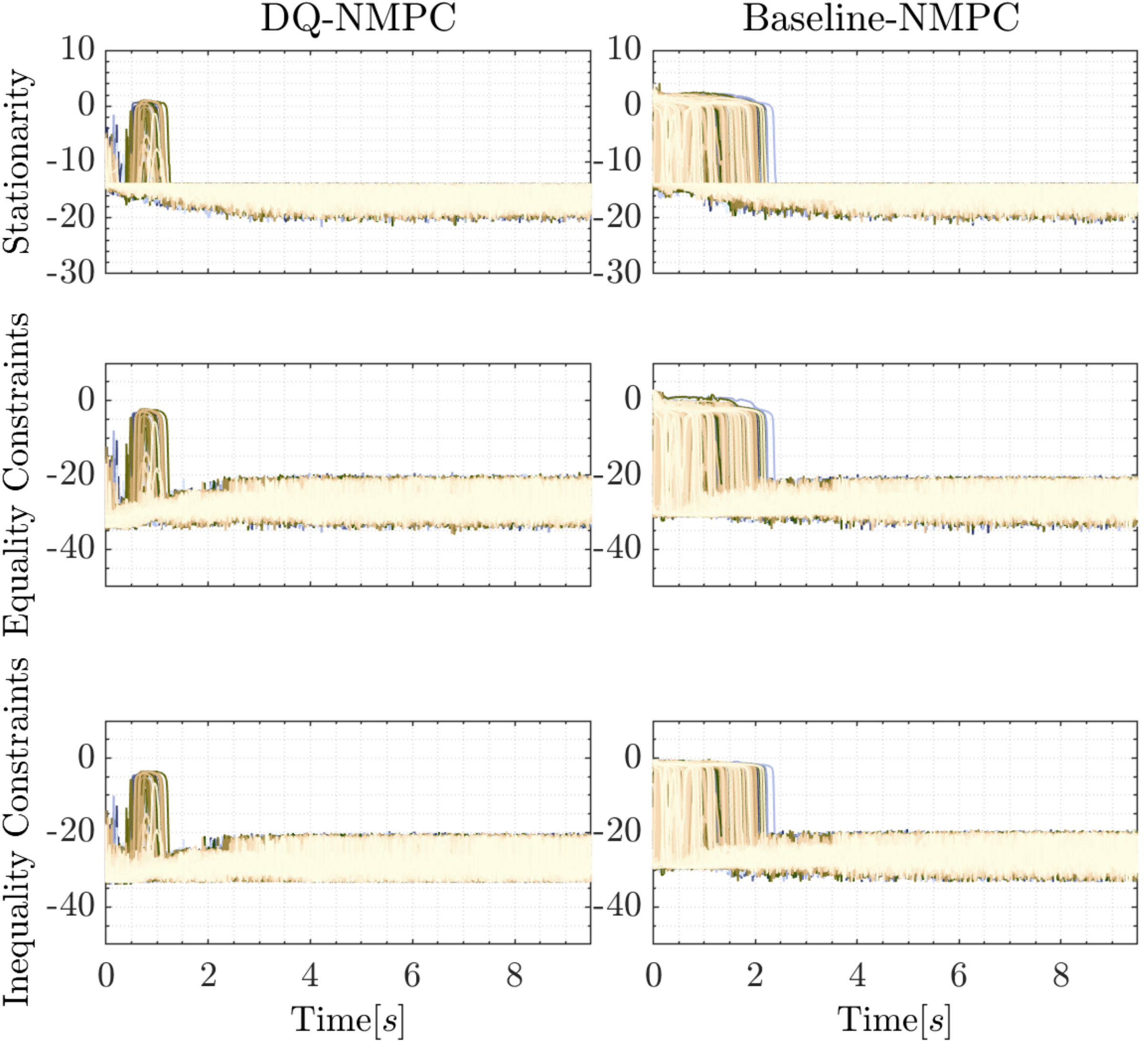}
    \caption{KKT condition residuals obtained from 600 simulations for pose regulation. The left column shows the results for DQ-NMPC, while the right column presents those for the baseline.}
    \label{fig:KKT}
    \vspace{-15pt}
\end{figure}

\subsubsection{Pose Regulation}

This section presents results related to the pose regulation task. We report the outcomes for both controllers by examining position and orientation metrics and evaluating how closely the optimization solutions approach optimality.
The desired attitude, represented as zero rotation quaternion and the desired position, represented as the origin are identical for both formulations. During these experiments, both NMPC methods were solved using Sequential Quadratic Programming (SQP), with HPIPM \cite{FRISON20206563} handling the internal QP subproblems, and the Gauss-Newton approximation, all using tools available in the ACADOS framework~\cite{verschueren2022acados}. A fixed prediction horizon of $1.5~\si{s}$ was used and the gain matrices were meticulously designed to ensure similar stabilization times.

\textbf{Control Performance}\label{control_performance}
To quantitatively evaluate the behavior of both controllers, we use two primary metrics: the position error, defined as $\| \mathbf{p}_{d} - \mathbf{p} \|$, and the orientation error, given by $\| \ln( \mathbf{r}_{d}^{*} \otimes \mathbf{r} ) \|$. 
 Performance metrics for both control formulations are presented in Fig.~\ref{fig:all_simulation_pose}, based on an evaluation of 600 initial positions and orientations randomly sampled from a uniform distribution. We observe that the position and orientation errors are influenced by large initial pose error, they affect the convergence of the baseline NMPC. On the other hand, the DQ-NMPC is able to maintain convergence across all initial conditions.

\begin{table*}[h!]
    \centering
    \caption{Performance metrics: comparison between dq-nmpc and baseline-nmpc under unmodeled dynamics and external disturbances.}
    \begin{tabular}{lcccccc}
        \toprule
        & \multicolumn{2}{c}{\bfseries Position RMSE [m]} & \multicolumn{2}{c}{\bfseries Orientation RMSE [rad]} & \multicolumn{2}{c}{\bfseries Rate of Change} \\
        & \multicolumn{2}{c}{mean $\pm$ STD} & \multicolumn{2}{c}{mean $\pm$ STD} & \multicolumn{2}{c}{$\Vert \Delta \mathbf{u}_f\Vert,\Vert\Delta \mathbf{u}_{\omega}\Vert$} \\
        \cmidrule(lr){2-3} \cmidrule(lr){4-5} \cmidrule(lr){6-7}
        & DQ-NMPC & Baseline-NMPC & DQ-NMPC & Baseline-NMPC & DQ-NMPC & Baseline-NMPC \\
        \midrule
        Disturbance-free ideal model & \textcolor{blue}{$0.039 \pm 0.005$} & $0.103 \pm 0.039$ & \textcolor{blue}{${0.083 \pm 0.011}$} & $0.269 \pm 0.121$ & \textcolor{blue}{$25.95$},~$14.17$ & $40.06$,~\textcolor{blue}{$13.26$} \\
        \midrule
          $+\%78$ Drag & \textcolor{blue}{$0.136 \pm 0.029$} & {$0.268 \pm 0.106$} & \textcolor{blue}{${0.182 \pm 0.036}$} & $0.322 \pm 0.082$ & \textcolor{blue}{$16.92,~6.21$} & $31.90,~11.12$ \\
        \midrule
        $+\%20$ Mass & $0.132 \pm 0.013$ & \textcolor{blue}{$0.128 \pm 0.026$} & \textcolor{blue}{${0.088 \pm 0.014}$} & $0.240\pm 0.075$ & \textcolor{blue}{$29.32$},~{$12.58$} & $47.71$,~\textcolor{blue}{$12.16$} \\
          \midrule
        $+\%20$ Inertia Matrix & \textcolor{blue}{$0.039 \pm 0.005$} & $0.081 \pm 0.021$ & \textcolor{blue}{${0.088 \pm 0.014}$} & $0.239 \pm 0.095$ & \textcolor{blue}{$23.75,~18.16$} & $42.39,~34.38$  \\
          \midrule
        $7.07~[N]$ External Force & $0.279 \pm 0.174$ & \textcolor{blue}{$0.240 \pm 0.119$} & \textcolor{blue}{${0.226 \pm 0.097}$} & $0.375 \pm 0.126$ & \textcolor{blue}{$30.53,~15.88$} & $46.54,~17.58$ \\
         \midrule
        $0.02~[Nm]$ External Moment & \textcolor{blue}{$0.152 \pm 0.012$} & $0.183 \pm 0.080$ & \textcolor{blue}{${0.157 \pm 0.052}$} & $0.651 \pm 0.560$ & \textcolor{blue}{$26.37,~16.10$} & $50.01,~21.56$ \\
        \bottomrule
    \end{tabular}
    \label{table:3}
\end{table*}

\textbf{Optimality Conditions}\label{optimality}
To assess the optimality of the solutions of the optimal control problems, we analyze the Karush-Kuhn-Tucker (KKT) condition numbers. 
Fig.~\ref{fig:KKT} shows that the DQ-NMPC consistently achieves better conditioned solutions near optimal, even in the presence of significant initial errors, commonly encountered at the beginning of the optimization process. This behavior improves the minimization of the cost function while satisfying both the equality and inequality constraints.
\begin{figure}[t]
    \centering
\includegraphics[width=\columnwidth]{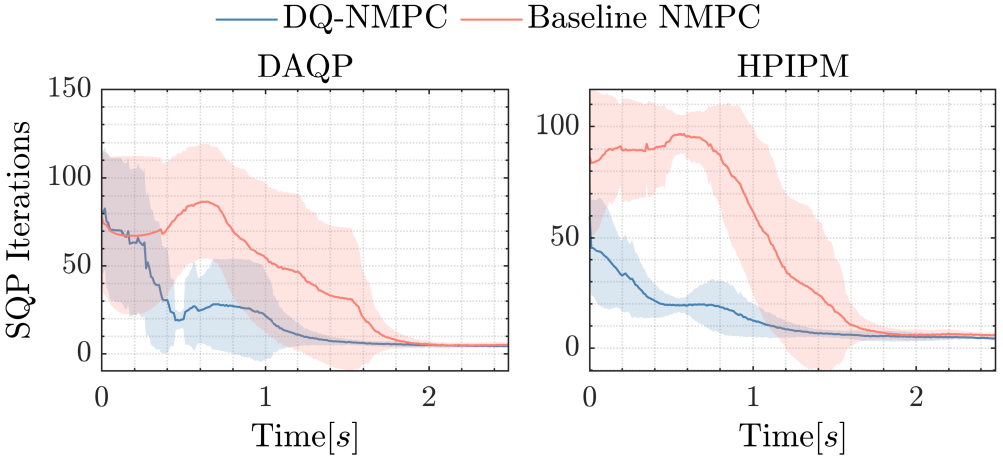}
    \caption{SQP iterations in the optimization process for pose regulation task considering HPIPM and DAQP as internal QP solvers under large initial pose errors.}
    \label{fig:SQP}
\vspace{-20pt}
\end{figure}

\textbf{Convergence Across QP solvers}\label{Convergence}
We report the results for both methods in a pose regulation task, evaluating 20 randomly sampled poses with only large pose errors. Two QP solvers were considered, HPIPM \cite{FRISON20206563} and DAQP \cite{9779534}, both commonly used in real-time applications. We present the mean and standard deviation of the number of iterations required by the SQP algorithm in Fig.~\ref{fig:SQP}. Despite the presence of large pose errors, the proposed DQ-NMPC consistently required fewer iterations to converge. We attribute this property to the unified modeling of the quadrotor dynamics on the dual-quaternion manifold and the use of a left-invariant pose error projected onto the Lie algebra of dual quaternions, which yields a well-scaled error representation and contributes to improved convergence of the SQP algorithm. 

\subsubsection{Robustness Study}

Previous simulations were conducted under ideal conditions, assuming perfect knowledge of the model and the absence of disturbances. However, this is not the reality in practical experiments. Therefore, this section studies the performance of DQ-NMPC and Baseline-NMPC under the following scenarios:

\begin{itemize}
\item \textit{Unmodeled Dynamics:} We include a model mismatch by considering drag, mass, and inertia coefficients in the quadrotor model.
\item \textit{External Disturbances} During the experiments, it is possible to introduce constant external forces, 
 and torques, which can significantly decrease the performance of both controllers.
\end{itemize}
The simulations were conducted in our custom environment, with both NMPC methods executed using SQP under the Real-Time Iteration (RTI) scheme. The internal QP subproblems were solved using HPIPM~\cite{FRISON20206563}, with a Gauss-Newton approximation, all implemented within the ACADOS framework~\cite{verschueren2022acados}.
These experiments are carried out considering feasible trajectories, using the circular $16$ and Lissajous trajectories, with a maximum velocity of $ V_{max} = 4.68~\si{m/s}$. 

\begin{figure}[t]
    \centering
\includegraphics[width=\columnwidth]{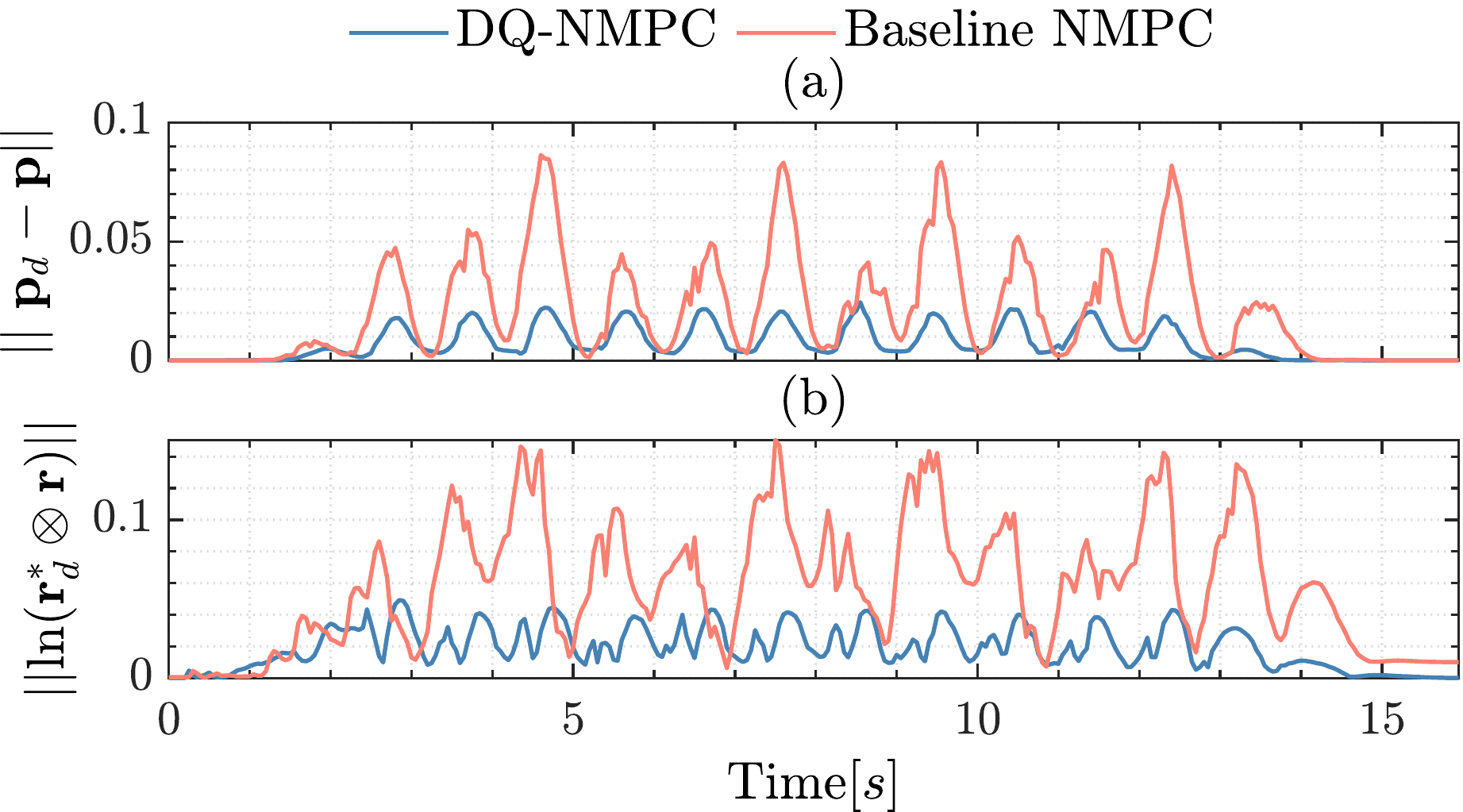}
    \caption{Tracking performance of a Lissajous trajectory with a maximum speed of up to $4.67~\text{m/s}$ (Disturbance-free and ideal model). \textbf{a}: Position error metric. \textbf{b} Orientation error metric}
    \label{fig:Simulation_complete}
    \vspace{-15pt}
\end{figure}

Table \ref{table:3} presents a comprehensive evaluation of the RMSE and the rate of change of control actions in various unmodeled dynamic conditions. The DQ-NMPC consistently outperforms the baseline in orientation tracking accuracy, achieving a significantly lower orientation RMSE in all scenarios. This advantage is particularly evident in experiments that involve inertia mismatches, external torques, and aerodynamic drag, where DQ-NMPC demonstrates superior performance. Furthermore, it produces smoother control inputs, which are critical for maintaining the stability of the system. Although the baseline NMPC achieves slightly lower position RMSE in certain situations, especially when the system is subject to mass variations and external forces, its performance decreases in orientation tracking and control smoothness. In general, the DQ-NMPC formulation provides more balanced behavior in a wide range of challenging conditions.
Fig. \ref{fig:Simulation_complete} illustrates the tracking results obtained for one of the trajectories in the simulation robustness study including the control error metrics. It is evident that DQ-MPC outperforms the baseline, particularly in reducing orientation errors.

\subsection{Real-World Experiments}\label{Real-World Experiments}

In this section, we present the real-world experiment results. The experiments are conducted in the Aerial-robot Control and Perception lab (ACP) flying space at Worcester
Polytechnic Institute. We used a customized quadrotor platform equipped with Nvidia Jetson Orin NX and PixRacer Autopilot board. The quadrotor flies in  a netted facility of dimension  $11\si{m} \times 4.5\si{m} \times 3.65\si{m}$ equipped with Vicon motion capture system, which provides the quadrotor state feedback at $240~\si{Hz}$. Both the proposed DQ-NMPC method and the baseline NMPC method are implemented in ROS2 using C++ on a Nvidia Jetson Orin NX at $190~\si{Hz}$. The optimization problem was solved in ACADOS~\cite{verschueren2022acados} using SQP with RTI scheme, with HPIPM \cite{FRISON20206563} handling the internal QP subproblems and a Gauss–Newton Hessian approximation over a fixed prediction horizon of $1.5~\si{s}$. To ensure a fair comparison, the weighting matrices for both controllers were tuned experimentally to achieve comparable performance under low-velocity conditions $4.7~\si{m/s}$. These parameters were kept fixed across all experiments and were not adjusted based on the specific trajectory or velocities.

\begin{table*}[t]
    \centering
    \caption{Performance metrics in real-world experiments: Position and orientation RMSE for different trajectories and velocities, along with the percentage improvement of DQ-NMPC over the baseline.}
    \begin{tabular}{llcccccc}
        \toprule
        \textbf{Trajectory} & \textbf{Max Velocity [m/s]} & \multicolumn{2}{c}{\textbf{Position RMSE [m]}} & \multicolumn{2}{c}{\textbf{Orientation RMSE [rad]}} & \multicolumn{2}{c}{\textbf{Error Reduction [\%]}} \\
        & & DQ-NMPC & Baseline & DQ-NMPC & Baseline & Position & Orientation \\
        \midrule
        Loop & 1.0 & \textcolor{blue}{$0.0135 \pm 0.0031$} & $0.0210 \pm 0.0042$ & \textcolor{blue}{$0.0115 \pm 0.0057$} & $0.0675 \pm 0.0927$ & 35.71 & 82.96 \\
         & 1.6 & \textcolor{blue}{$0.0163 \pm 0.0046$} & $0.0246 \pm 0.0058$ & \textcolor{blue}{$0.0277 \pm 0.0141$} & $0.0768 \pm 0.0439$ & 33.73 & 63.80 \\
         & 1.9 & \textcolor{blue}{$0.0146 \pm 0.0054$} & $0.0298 \pm 0.0075$ & \textcolor{blue}{$0.0271 \pm 0.0196$} & $0.1034 \pm 0.0580$ & 51.00 & 73.80 \\
        \midrule
        Lissajous & 4.7 & {$0.0326 \pm 0.0187$} & \textcolor{blue}{$0.0324 \pm 0.0110$} & \textcolor{blue}{$0.0465 \pm 0.0262$} & $0.0811 \pm 0.0348$ & -0.61 & 42.66 \\
         & 5.3 & \textcolor{blue}{$0.0384 \pm 0.0238$} & $0.0386 \pm  0.0131$ & \textcolor{blue}{$0.0506 \pm 0.0316$} & $0.0856 \pm  0.0335$ & 0.51 & 40.88 \\
         & 6.1 & \textcolor{blue}{$0.0443 \pm 0.0280$} & $0.0581 \pm 0.0347$ & \textcolor{blue}{$0.0549 \pm 0.0382$} & $0.0864 \pm 0.0478$ & 23.74 & 36.45 \\
         & 7.5 & \textcolor{blue}{$0.0577 \pm 0.0413$} & $0.1314 \pm 0.0987$ & \textcolor{blue}{$0.0568 \pm 0.0432$} & $0.1314 \pm 0.0670$ & 56.11 & 56.77 \\
        \bottomrule
    \end{tabular}
    \label{tab:real_comparative}
\end{table*}
Experimental evaluation involves trajectory tracking experiments with circular and Lissajous trajectories, achieving velocities of up to $7.5~\si{m/s}$. Table \ref{tab:real_comparative} clearly highlights the performance of the proposed DQ-NMPC. DQ-NMPC demonstrates a substantial reduction in orientation tracking errors, achieving improvements of up to $56.77\%$ relative to the baseline formulation considering velocities up to $7.5~\si{m/s}$. Furthermore, it consistently maintains a lower position RMSE across the evaluated range of velocities. In particular, the performance gap widens significantly at higher velocities, where the baseline formulation exhibits a noticeable degradation in orientation tracking accuracy. These findings underscore the effectiveness of the DQ-NMPC, validating its advantages for high-speed scenarios. 

\begin{figure}[t]
    \centering
\includegraphics[width=\columnwidth]{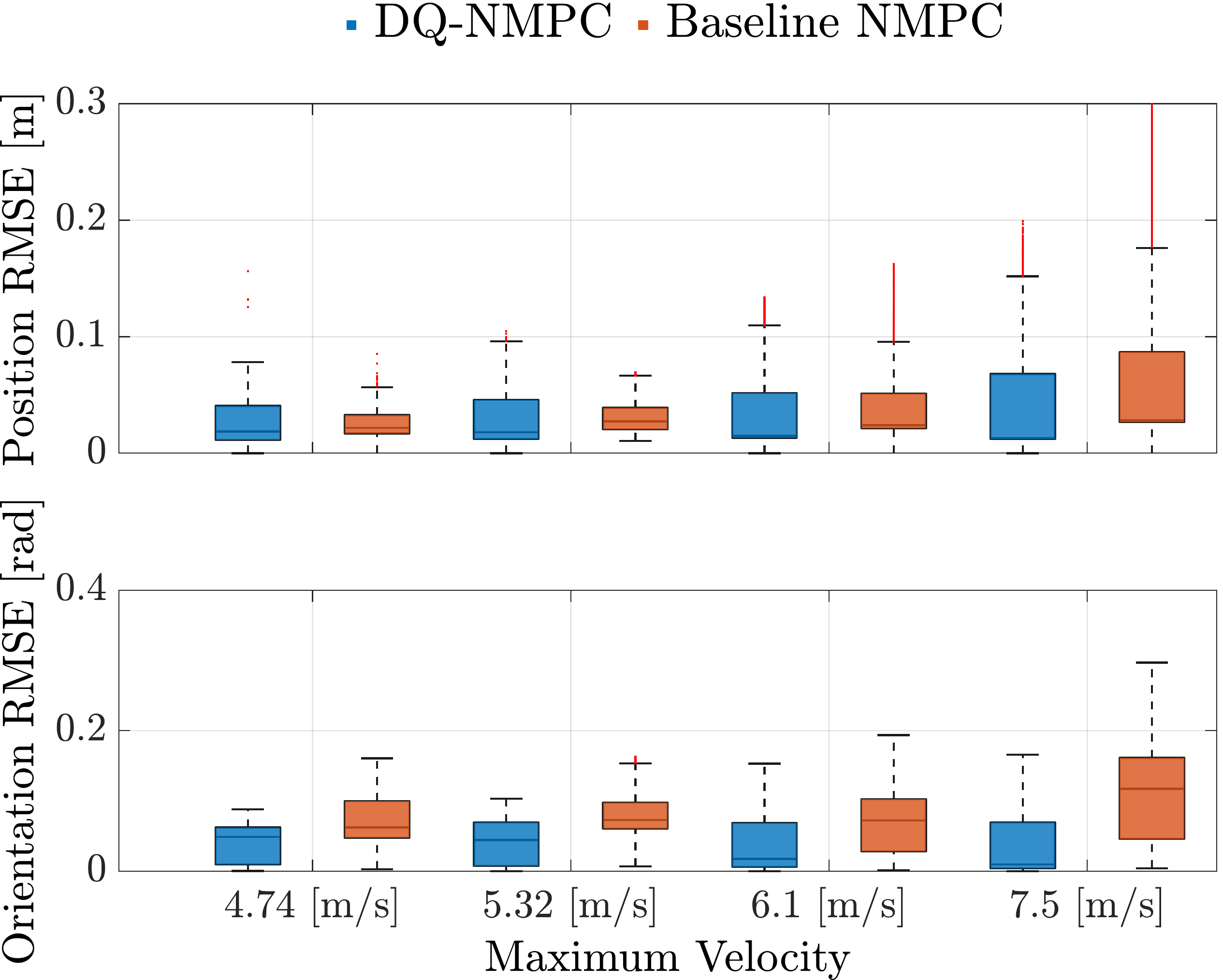}
    \caption{Box plot of tracking error for position and orientation considering a Lissajous in real-world experiments.}
    \label{fig:box_liss_real}
\end{figure}

Fig.~\ref{fig:box_liss_real} presents a comparison of boxplots for DQ-NMPC and the baseline NMPC controller when tracking Lissajous trajectories at varying reference maximum velocities. In terms of position RMSE, both methods exhibit similar performance, with DQ-NMPC showing a slightly lower average error. However, in orientation tracking, DQ-NMPC clearly outperforms the baseline. Notably, as the reference velocity increases, the baseline controller's performance degrades significantly, while DQ-NMPC maintains consistent accuracy. 

We attribute the tracking performance of the proposed DQ-NMPC to the left-invariant pose error on the dual-quaternion group, which includes orientation and translation in the desired frame.
Additionally, since the pose error belongs to a dual-quaternion group, the error is projected onto its Lie algebra, enabling the design of a quadratic cost function that provides a minimal and symmetric representation. Unlike conventional metrics that lose sensitivity under large orientation deviations~\cite{9794477, 5717652}, our cost maintains sensitivity throughout large orientation errors, ensuring fast convergence of rotational dynamics and accurate trajectory tracking.

\begin{figure}[t]
    \centering
\includegraphics[width=\columnwidth]{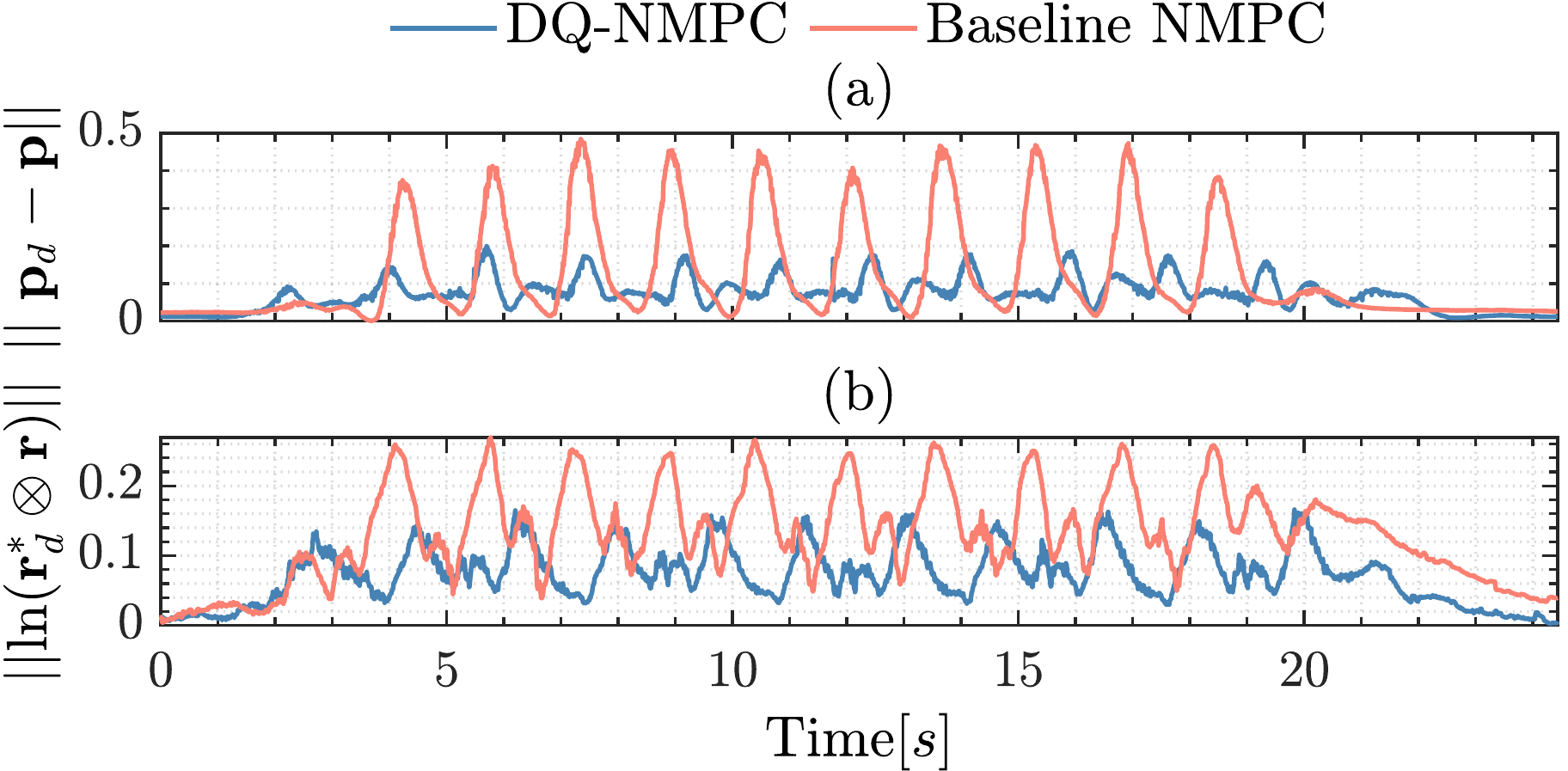}
    \caption{Tracking performance of a Lissajous trajectory with a maximum speed of up to $7.5~\text{m/s}$. \textbf{a}: Position error metric. \textbf{b}:  Orientation error metric.}
    \label{fig:liss_real_experiment}
    \vspace{-15pt}
\end{figure}

The corresponding performance metrics, including position and orientation errors, are shown in Fig.~\ref{fig:liss_real_experiment}a and~\ref{fig:liss_real_experiment}b, respectively. The results demonstrate that the DQ-NMPC controller achieves superior performance while maintaining consistent pose tracking accuracy throughout the experiment.

\begin{figure}[t]
    \centering
\includegraphics[width=\columnwidth]{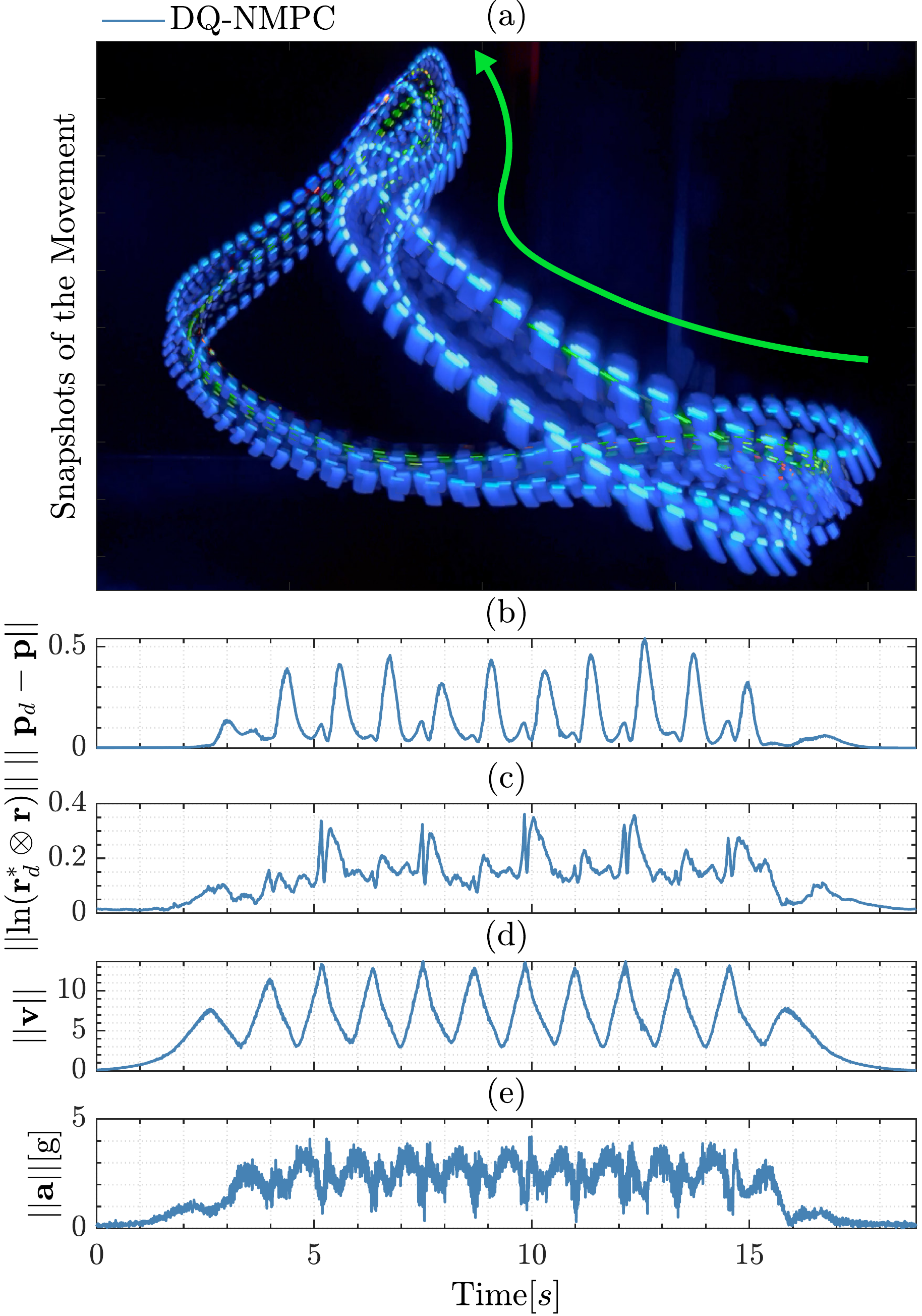}
    \caption{Tracking performance of a Lissajous trajectory with a maximum speed of up to $13.66~\text{m/s}$ and accelerations up to $4.2~\si{g}$. \textbf{a}: Movement of the quadrotor using the DQ-NMPC. \textbf{b}: Position error metric. \textbf{c}:  Orientation error metric. \textbf{d}:  Magnitude of the linear velocity during the experiment. \textbf{e}:  Magnitude of the accelerations during the experiment.}
    \label{fig:liss_real_experiment_fast}
    \vspace{-15pt}
\end{figure}

Finally, we push the limits of the proposed DQ-NMPC framework by conducting experiments with Lissajous trajectories at speeds of up to $13.66~\si{m/s}$ and accelerations up to $4.2~\si{g}$. Snapshots of the quadrotor executing this high-speed trajectory are shown in Fig.~\ref{fig:liss_real_experiment_fast}a. The corresponding performance metrics are presented in Fig.~\ref{fig:liss_real_experiment_fast}b and Fig.~\ref{fig:liss_real_experiment_fast}c, respectively. The velocity and acceleration profiles achieved by the quadrotor during the experiment are illustrated in Fig.~\ref{fig:liss_real_experiment_fast}d and Fig.~\ref{fig:liss_real_experiment_fast}e, respectively. No baseline NMPC results are reported for this scenario, as the baseline controller was unable to compute a feasible solution at such speeds. 

\section{Conclusion}\label{Conclusion}
This work introduces the DQ-NMPC framework for quadrotor flight. By leveraging unit dual-quaternions, the proposed approach provides a unified
and compact representation of quadrotor dynamics.  The proposed dual-quaternion-based cost function also simultaneously penalizes both
rotation and translation errors. 
Extensive simulations and real-world experiments demonstrate the effectiveness of the proposed method, showing an improved convergence of the optimization problem and enhanced tracking performance compared to a conventional baseline NMPC method. Future work will also explore the application of the DQ-NMPC framework to other robotic platforms, such as aerial manipulators,  fully actuated multirotor vehicles, and legged robots, as well as its integration with learning-based methods to enhance adaptability and performance in dynamic and uncertain environments.

\bibliography{references.bib}{}
\bibliographystyle{IEEEtran}
\end{document}